# Optimal Synthesis of Finite State Machines with Universal Gates using Evolutionary Algorithm

[1]Noor Ullah, [2] Khawaja M. Yahya, [3]Irfan Ahmed

[1,2,3] Department of Electrical Engineering

University of Engineering and Technology, Peshawar, Pakistan.

E-mail:  [1]noorullah51@gmail.com, [2]yahyakm@yahoo.com, [3]irfanahmed@nwfpuet.edu.pk

## ABSTRACT

This work presents an optimization method for the synthesis of finite state machines. The focus is on the reduction in the on-chip area and the cost of the circuit. A list of finite state machines from MCNC91 benchmark circuits have been evolved using Cartesian Genetic Programming. On the average, almost *30%* of reduction in the total number of gates has been achieved. The effects of some parameters on the evolutionary process have also been discussed in the paper.

## 1. INTRODUCTION

Circuit size and cost are among the main issues in digital circuit design these days. Finite State Machines (FSMs) are considered as the heart of sequential systems. FSMs are generally referred to the two models of sequential digital circuits namely Mealy and Moore models. Mealy model describes the output of a system as a function of both the input and current state.  Moore model expresses it in terms of the current state only. These sequential systems consist of combinational logic circuits, which are connected to the storage elements making feedback path. Designing of FSMs involves seven crucial steps [1]. One of the most vital steps is to obtain optimal state equations, which include input and output variables. This is an indispensable step towards an efficient, small and cost effective hardware design.

This paper focuses on the objective to obtain an optimal combinational logic circuit for the FSM. The primary goal is to reduce the number of gates as much as possible. This leads to a reduction in total number of MOSFETs, which saves the on-chip area of FSMs and reduces the cost of circuit as well.

Many researchers have worked on the optimization of FSMs using different techniques. A symbolic description of FSMs has been considered for logic minimization in PLA based machines [2]. *S. Devadas* introduced algorithms using migration and utilization of don't-care sequences [3]. Cellular automata based synthesis scheme for sequential circuits has been described [4]. Different genetic algorithms have been proposed in the area [5-9]. Heuristic algorithms for two level logic and two-hot encoding have been introduced [10],[11]. R. *S. Shelar et. al.,* have decomposed FSMs into two interactive machines [12]. Mean Field Annealing based solution of graph-embedding problem has been given [13]. *L. Yuan et. al.,* have devised a state splitting technique for FSMs [14]. The aforementioned research work is mainly aimed at one of the two goals or both: First to reduce the number of states that describe the behavior of FSMs. Secondly, to encode the states with such binary sequences that the switching between the states is minimized. In any case the purpose is a reduction in total area either as a primary objective or secondary. Three MCNC benchmark FSMs have also been evolved using a Genetic Programming (GP) [15]. However the above research evolves FSMs using many types of logic gates. *Shanti et. al.,* have presented the evolution of asynchronous sequential circuits using developmental Cartesian Genetic Programming [16]. The research was carried out to evolve the combinational part for each memory element individually.

In this paper, the evolution of combinational logic part of FSMs has been proposed using Cartesian Genetic Programming (CGP). The circuits are evolved with universal gates (NAND and NOR) only. The evolution encompasses all the state elements and the outputs of the system together in the same program. The benefit of the combined evolution is that many redundant nodes are removed. This results in very compact circuit architecture. The rest of the paper is organized in the following manner. Section 2 gives an overview of CGP. Section 3 describes the detailed experimental setup. Section 4 is about the simulation process and results of the experiments. Section 5 concludes the paper.

## 2. CARTESIAN GENETIC PROGRAMMING

CGP is a variant of GP, which was invented for the evolution of digital circuits by *Miller* and *Thompson* [17]. In CGP programs are represented as directed acyclic graphs in contrast to the conventional tree-based GP. This allows the indirect reuse of the nodes. In start the CGP graphs were represented by two-dimensional grid of nodes. Any number of rows, columns and level backs could be chosen by the user, creating a number of different topologies. Later work showed that a special case, having a single row with level backs equal to the number of columns was more effective [18].

A list of integers called genes represents a fixed-length genotype in CGP. It encodes all the incoming and outgoing connections and the function of each node. The decoded genotype is called phenotype. Its size can vary from zero nodes to the maximum number of nodes in the genotype. This is due to the fact that all the nodes in genotype may not be used in the phenotype. Such nodes and their genes that have no influence on phenotype are called non-coding. They have a neutral effect on the fitness of genotype often called as neutrality, which is discussed in detail in [17]. Each computational node in the genotype represents a function from a user defined list in a function look-up table and is encoded by two types of genes:

i. The address of a node-function in the function look-up table is called a function gene. It decides the operation of the node and is always the first gene in the node.
ii. Connection genes are determined by the arity of any function. They encode the input connections of the nodes and are basically indices in an array.

In the special case topology described above, a program input or any previous node's output can become the input of the nodes in a feed forward manner. Absolute values are assigned to program inputs from 0 to $n_i-1$ where $n_i$ is the number of inputs. The nodes' outputs are also ordered in a sequential manner from 0 to $ni+m-1$, where $m$ is the maximum number of nodes. At the end of the genotype, as many integers as the required number of outputs are added representing the outputs, where Each integer represents the address of the node from where the output is taken. The general form of Miller's CGP is shown in figure 1.

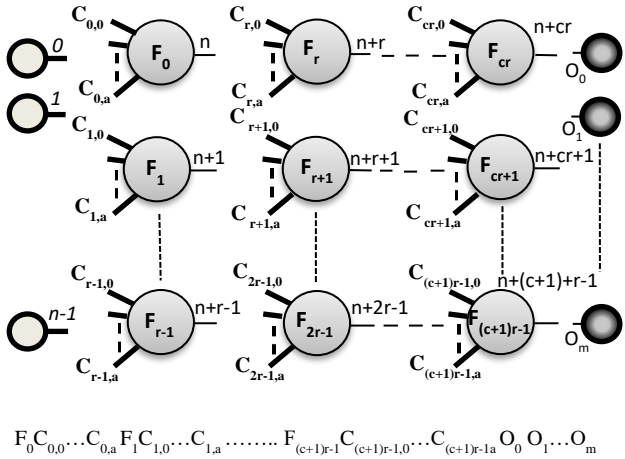

$F_0 C_{0,0}…C_{0,a} F_1 C_{1,0}…C_{1,a} ……… F_{(c+1)r-1} C_{(c+1)r-1,0}…C_{(c+1)r-1a} O_0 O_1…O_m$

**Fig.1.** General Form of Miller's CGP, where $n_i$, $n_r$ and $n_c$ and $a$ represent number of inputs, number of rows, number of columns and arity respectively [19]

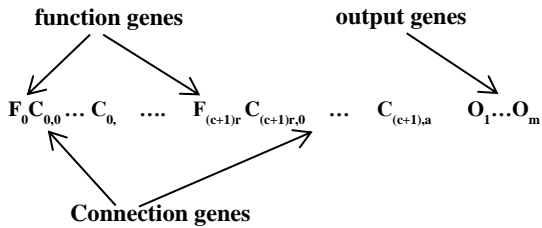

**Fig.2.** CGP Genotype [19]

In CGP, some constraints must be obeyed at the initialization or mutation of genotype. The function gene's allele $f_i$ for a total number of functions $n_f$ must obey the following relation:

$$0 \leq f_i \leq n_i \quad (1)$$

The connection genes' alleles $C_{ij}$ for all nodes in column $j$ must follow the given relations, where $l$ is the value of level-backs:

$$n_i + (j-l)n_r \leq C_{ij} \leq n_i + jn_r, \text{ if } j \geq l \quad (2)$$
$$0 \leq C_{ij} \leq n_i + jn_r, \text{ if } j < l \quad (3)$$

In CGP, a point mutation operator is used, in which a randomly chosen gene's value is changed with another valid random value. A valid value for a function gene is any index in the function look-up table. For an input gene it is the index of any previous node's output or any program input. For an output gene it can take the output index of any node or program input. Mutation rate, $\mu_r$ is a user defined value (a percentage of total number of genes), which gives the number of mutations per application of mutation. All the off-springs go through the mutation process.

In any CGP program, a fitness criterion must be set, based on which the decision of when to stop the evolution process is taken. This is determined by the magnitude of error between the evolved output and the desired output. In order to achieve *100%* accuracy, the magnitude of error must be equal to *0*. During the process among the parent and all the off-springs, the one with the least error is considered as the fittest and is promoted as a parent for the next iteration. If a parent and an off-spring have the same least error then the off-spring is considered as the fittest. If all of them have the same magnitude of error then any one of the randomly chosen off-springs is considered as the fittest. CGP evolution is performed normally with a simplified form of *1+λ* evolutionary algorithm, where *λ* is the number of off-springs. CGP decoding is done from the output to the input to yield the phenotype. The non-coding nodes are ignored in this process.

## 3. EXPERIMENTAL SETUP

A one dimensional CGP graph is used for all the experiments, which consists of a single row and m number of columns and level-backs. The input array in the program contains all the possible combinations of the system inputs and the current state values of all the flip flops of FSM. An array stores the desired outputs for all the combinations of input array. These outputs are comprised of the system outputs and the next state output values of all the flip flops. Another array contains the outputs that are created per iteration by the CGP program, for all combinations of input array. A simplified form of *1+λ* evolutionary algorithm is used for the evolution purpose. Two values of *λ* (4 and 8) are used to evolve each FSM separately. The maximum number of nodes in the program varies according to the requirement of each FSM. The function lookup table contains only NAND and NOR functions (which make the universal gates), taking only two inputs each. The root mean square error between the desired outputs and the CGP evolved outputs, decides the fitness of the parent and off-springs. To achieve *100%* accuracy in the design, the least error must become zero. Point mutation operator creates the randomly mutated off-springs. A mutation rate ranging from *3%* to *10%* is used, which depends on the size of each circuit. The CGP code is written in C++.

## 4. SIMULATION AND RESULTS

The CGP evolved circuit design is compared with the espresso based architecture of six MCNC91 benchmark FSMs. The same comparison is done for four custom made Moore type sequence detectors also. First, the state transition tables are fed into the espresso based software, *logic Friday* for logic minimization. The minimized logic equations are mapped into universal gates in *Logic Friday*. The obtained combinational circuit is connected with D flip flops, constructing complete

FSM. To get the optimized design, the same logic tables are used to evolve the circuit in CGP. The evolved circuits have been tested for all combinations of inputs and yield accurate outputs. The results are shown in Tables 1 and 2.

**TABLE 1:** Comparison b/w conventional and CGP designed MCNC91 benchmark FSM circuits

| Name | No. of Gates | | Percentage Reduction | No. of States | No. of Inputs | No. of Outputs |
|---|---|---|---|---|---|---|
| | *Espresso* | *CGP* | | | | |
| dk27 | 23 | 18 | 21.73 | 7 | 1 | 2 |
| Lion9 | 25 | 19 | 24 | 9 | 2 | 1 |
| S8 | 31 | 22 | 29.03 | 5 | 4 | 1 |
| beecount | 38 | 26 | 31.57 | 7 | 3 | 4 |
| bbara | 62 | 43 | 30.64 | 10 | 4 | 2 |
| dk14 | 124 | 79 | 36.29 | 7 | 3 | 5 |

**TABLE 2**: Comparison b/w conventional and CGP designed custom made Moore type sequence detector FSMs

| Name | No.of Gates | | Percentage Reduction | No. of States |
|---|---|---|---|---|
| | *Espresso* | *CGP* | | |
| 10101 | 23 | 19 | 17.39 | 6 |
| 0001000 | 27 | 18 | 33.33 | 8 |
| 01100110 | 30 | 20 | 33.33 | 9 |
| 12-0s-then-1 | 42 | 20 | 52.38 | 14 |

Almost *30%* reduction in number of gates has been achieved in CGP evolved MCNC91 FSMs as compared with the conventional espresso based design. Similarly 34% reduction in custom made Moore type sequence detectors has been achieved. This is a significant amount of reduction in the total number of MOSFETS used to construct the FSMs. So a lot of on-chip area can be saved using CGP evolved FSMs. Also smaller design reduces the total cost of the circuit especially in larger FSMs. From the above data it is also evident that the reduction in total number of gates using CGP is independent of the number of states of FSMs. To demonstration the above results, the circuit diagrams of *dk27* for the espresso based design and CGP evolved circuit are shown in Fig. 3a and b.

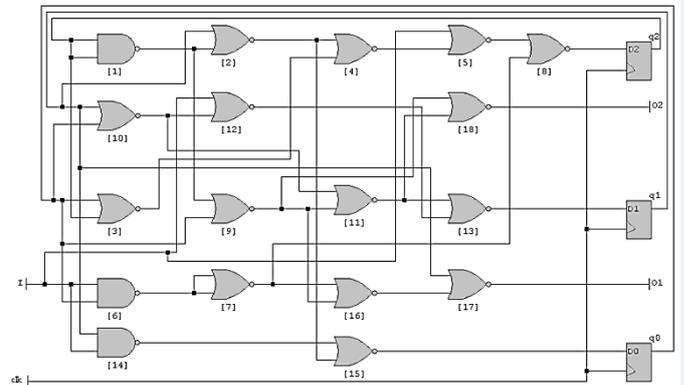

**Fig. 3b.** dk27 circuit with universal gates CGP based design

As mentioned in *section 3* that two different values of $\lambda$ are used for CGP evolution, different circuit architectures have been evolved for each value. The results are shown in Table 3 and Plots 1 and 2.

**TABLE 3:** Comparision of the effects $\lambda =4$ on circuit size and simulation time with $\lambda =8$

| Name | m | Nodes used | | No. of Generations | | μr |
|---|---|---|---|---|---|---|
| | | $\lambda=4$ | $\lambda=8$ | $\lambda=4$ | $\lambda=8$ | |
| dk27 | 25 | 21 | 18 | 199855 | 372007 | 10 |
| s8 | 25 | 24 | 22 | 367382 | 1634471 | 10 |
| lion9 | 25 | 19 | 18 | 2747861 | 3200766 | 10 |
| beecount | 55 | 30 | 28 | 192728 | 985972 | 3 |

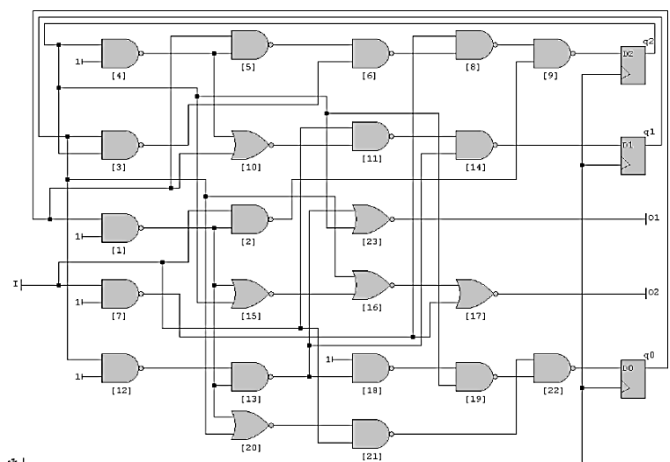

**Fig. 3a.** *dk27* circuit with universal gates espresso based design

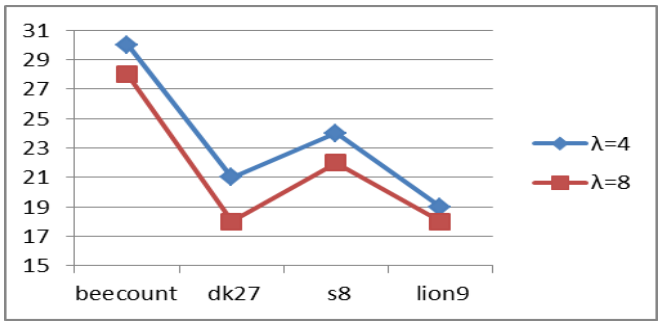

**Plot. 1**. Number of nodes used in phenotype vs λ

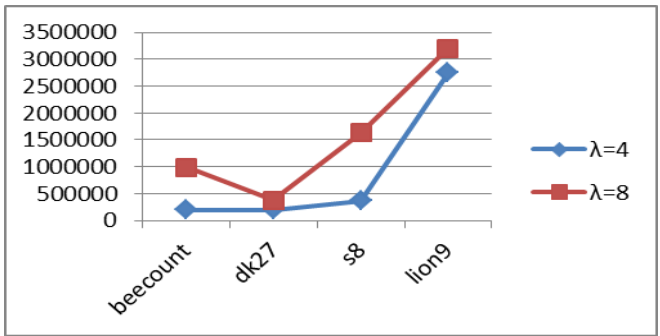

**Plot. 2.** Number of generation vs λ

In the above results, it is observed that the evolution process is much faster with a smaller value of *λ*. *λ=4* needs lesser number of generations to converge into the desired circuit than *λ=8*. On the other hand, the later value of *λ* evolves circuits with lesser number of gates than the former. So a simplified version of *(1+8)* evolutionary algorithm is a better choice for more compact circuits, which is the primary goal of the paper.

It is also observed that for most of the time, a smaller value of m in the genotype creates much smaller circuits. However there is a bound on the least value of *m*, below which simulation will take forever to converge into a particular FSM. As an example *bbara* can be evolved with *m=89* and *m=64*, where the first case uses *53* nodes to construct the circuit while the second needs only *43* nodes. Another parameter of CGP that plays a great role in the simulation time and up to some extent on the circuit size is mutation rate. A smaller value of $\mu_r$ evolves the FSMs much faster but with a slightly bigger size. So in case of smaller FSMs like *dk27* and *s8* a mutation rate of *10%* is used while *3%* in bigger circuits like *bbara*. Also finding suitable value of m could get extremely difficult in bigger circuits with higher mutation rate.

The circuit diagrams for other evolved MCNC91 benchmark FSMs used in the research are given below:

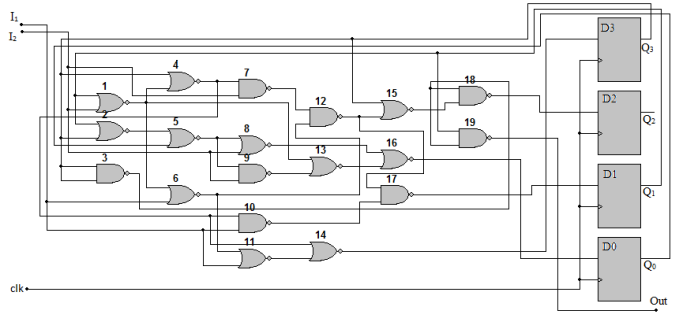

**Fig. 4.** *lion9* circuit with universal gates CGP based design

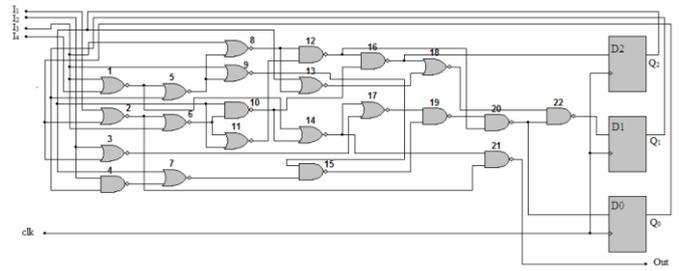

**Fig. 5.** *s8* circuit with universal gates CGP based design

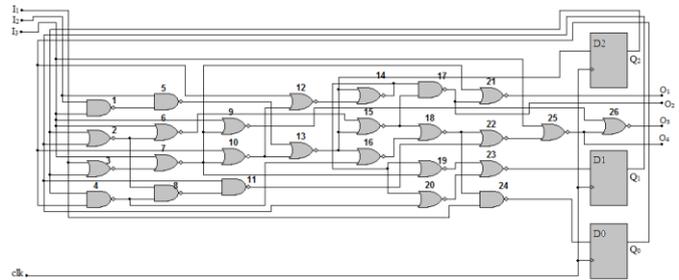

**Fig. 6.** *beecount* circuit with universal gates CGP based design

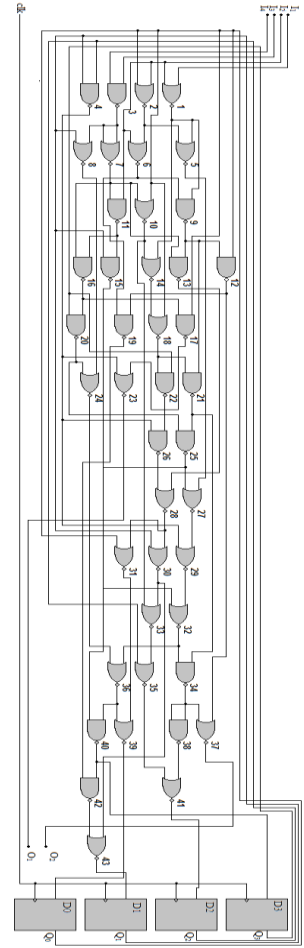

**Fig. 7.** *bbara* circuit with universal gates CGP based design

## 5. CONCLUSION

CGP based design shows that a significant reduction in the size of FSMs can be achieved as compared with the conventional K-Map or espresso based design hence saving a lot of on-chip area and money which is the need of the day in digital electronics industry. To achieve this goal, suitable selection of certain CGP parameters is of great importance determining the trade-off with simulation time.

In future this work can be extended to design and optimize more complex sequential circuits in terms of power, cost, size and propagation delay using either the proposed functions or other Boolean functions. Different CGP parameters can also be evaluated to achieve the required goal.

## ACKNOWLEDGEMENTS

The authors thank Engr. Fahad Ullah (RA and Ph.D. Scholar at CS dept. CSU, Colorado, USA) for his help in understanding the key concepts of CGP.